\documentclass{article}
\usepackage{spconf}
\usepackage{cite}
\usepackage[english]{babel}
\usepackage{graphicx}
\usepackage{epstopdf}
\graphicspath{{eps/}}
\usepackage{upgreek}
\DeclareGraphicsExtensions{.eps}
\usepackage[cmex10]{amsmath}
\usepackage{amsfonts}
\usepackage{amssymb}
\usepackage{enumerate}
\usepackage{subcaption}
\usepackage{url}
\usepackage{booktabs}

\usepackage{multirow}
\usepackage{scrextend}
\usepackage{tabularx}
\usepackage{times}
\usepackage{verbatim}
\usepackage{color}
\usepackage{amssymb}
\usepackage{algorithm}
\usepackage{algpseudocode}
\usepackage{hyperref}
\usepackage{xcolor}

\usepackage{fixltx2e}
\usepackage{mathtools}
\usepackage{textcomp}

\title{Region Proposal Networks with Contextual Selective Attention for Real-Time Organ Detection}
\name{Awais Mansoor\textsuperscript{a}\thanks{Address correspondences to: awais.mansoor@gmail.com}, Antonio R. Porras\textsuperscript{a}, Marius George Linguraru\textsuperscript{a,b}}
\address{\textsuperscript{a}Sheikh Zayed Institute of Pediatric and Surgical Innovation,\\ Children\textquotesingle s National Health System, Washington D.C., USA.\\
\textsuperscript{b}The School of Medicine and Health Sciences, George Washington University, Washington D.C., USA.}
\begin{document}
%\ninept
%
\maketitle
\begin{abstract}
State-of-the-art methods for object detection use region proposal networks (RPN) to hypothesize object location. These networks simultaneously predicts object bounding boxes and \emph{objectness} scores at each location in the image. Unlike natural images for which RPN algorithms were originally designed, most medical images are acquired following standard protocols, thus organs in the image are typically at a similar location and possess similar geometrical characteristics (e.g. scale, aspect-ratio, etc.). Therefore, medical image acquisition protocols hold critical localization and geometric information that can be incorporated for faster and more accurate detection. This paper presents a novel attention mechanism for the detection of organs by incorporating imaging protocol information. Our novel selective attention approach (i) effectively shrinks the search space inside the feature map, (ii) appends useful localization information to the hypothesized proposal for the detection architecture to learn where to look for each organ, and (iii) modifies the pyramid of regression references in the RPN by incorporating organ- and modality-specific information, which results in additional time reduction. We evaluated the proposed framework on a dataset of 768 chest X-ray images obtained from a diverse set of sources. Our results demonstrate superior performance for the detection of the lung field compared to the state-of-the-art, both in terms of detection accuracy, demonstrating an improvement of $>7\%$ in Dice score, and reduced processing time by $27.53\%$ due to fewer hypotheses.
\end{abstract}
\begin{keywords}
Organ detection, selective attention, region proposal network, lung field detection, chest radiograph.
\end{keywords}

\section{Introduction}
Organ detection is critical step in various medical image analysis applications including segmentation, semantic navigation, query processing, etc. The performance of these applications is contingent upon fast and accurate localization of the organ of interest. Moreover, with the application of big data technologies on the rise in the field of medical imaging, these technologies are even more essential to provide better patient care, create population-specific atlases, and curate data accurately for artificial intelligence algorithms. However, challenges in fast and accurate organ detection continue to be a bottleneck in the development of real-time and accurate medical imaging applications.

Traditional methods for organ detection are based primarily on a sliding window approach to generate hypotheses for the organ location. Then a classifier assigns labels to the hypotheses. The seminal real-time object detection algorithm proposed by Viola and Jones \cite{viola2001rapid} follows this approach using a cascade of Adaboost classifiers. In addition to impacting various computer vision applications, this method has also been adopted in a number of medical imaging applications. However, in medical imaging regression-based solutions are a more feasible route for organ detection than exhaustive search. This is due to the facts that: (i) exhaustive search is unnecessary when detecting anatomy because medical images offer strong contextual information, and (ii) the size and resolution of normative medical and biomedical images make exhaustive search prohibitively expensive. In addition, variations in orientation and scale increase the computational complexity exponentially. 

Several approaches to incorporate contextual information for efficient organ detection have been proposed in the literature. For instance, Zhou \emph{et al.} presented a method based on boosting ridge regression to detect and localize the left ventricle in cardiac ultrasound \cite{zhou2007boosting}. Pauly \emph{et al.} \cite{pauly2011fast} used supervised regression from 3D local binary pattern descriptors for organ detection in multichannel magnetic resonance (MR) Dixon sequences. Zhang \emph{et al.} \cite{zheng2014marginal} proposed marginal space learning (MSL), which breaks down the complexity of learning similarity transformation from the image space to the projection space. Although there has been an increasing interest in applying deep-learning methods for organ detection from medical images, the state-of-the-art techniques use either exhaustive search mechanism \cite{girshick2015fast, ren2015faster} or data pre-processing prior to neural network to incorporate contextual information \cite{ghesu2016marginal}. To our best knowledge, there are no works incorporating contextual information to deep neural networks for fast object and organ detection.

In this paper, we demonstrate how contextual information from image acquisition protocols about the organ location can be incorporated into state-of-the-art neural network-based detection mechanisms, such as RPN, resulting in faster and more accurate organ detection. The main contributions of the paper are: (i) we propose a reduced-size search space inside the convolutional feature map for proposal generation, (ii) we include useful prior information about the organ localization to the detection architecture so it can learn where to look for each organ, and (iii) we modify the pyramid of regression references in the RPN by incorporating organ- and modality-specific information, which results in additional time reduction and improved detection accuracy. We evaluate our proposed framework on the detection of the lung field from a dataset of 668 chest X-ray images obtained from diverse sources and compared it with state-of-the-art.
\vspace{-0.1in}
\section{Methods}
\begin{figure}
\begin{subfigure}[b]{0.45\textwidth}
\includegraphics[width=\textwidth]{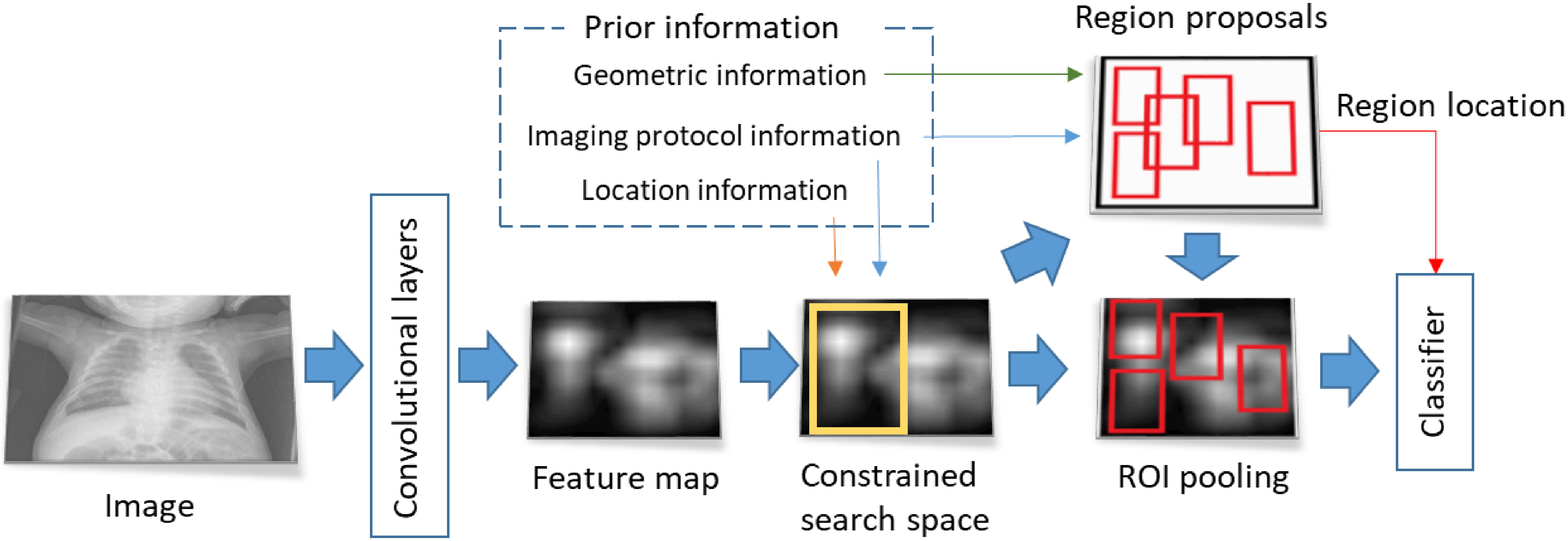}
\caption{}
\label{fig:rpn_ours}
\end{subfigure}
\begin{subfigure}[b]{0.45\textwidth}
\includegraphics[width=\textwidth]{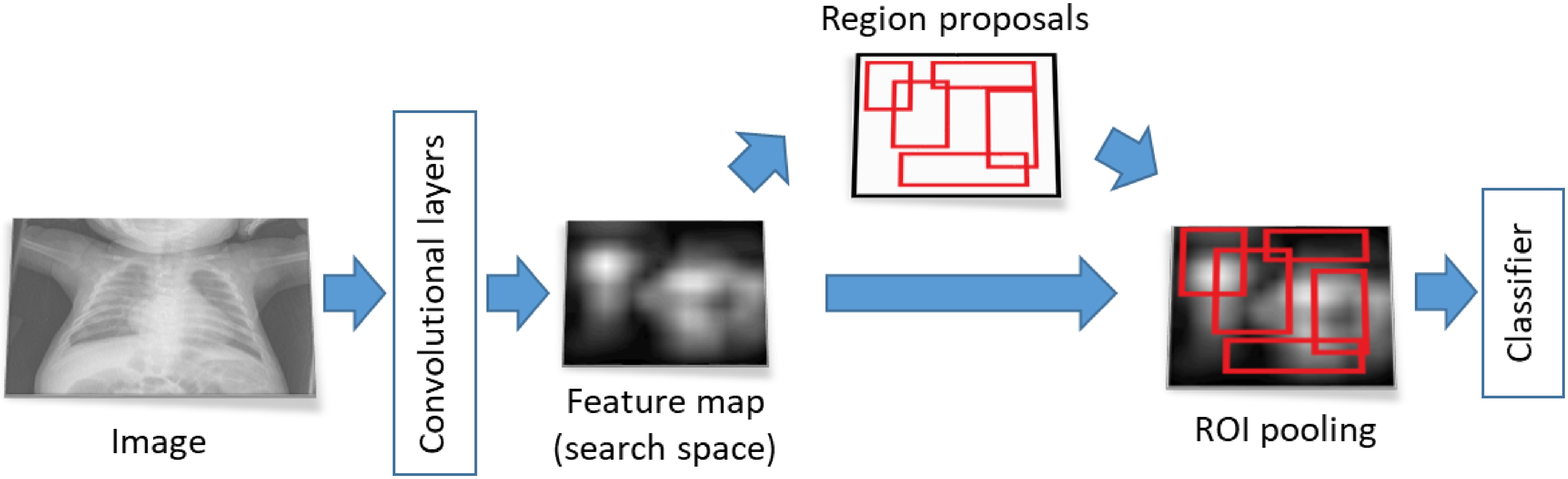}
\caption{}
\label{fig:rpn_original}
\end{subfigure}
\caption{\footnotesize{Comparative illustration of the proposed architecture of the network based on context aware selective attention (a), with the state-of-the-art: Faster R-CNN \cite{ren2015faster} (b).}}
\label{fig:rpn}
\end{figure}

Fig. \ref{fig:rpn_ours} provides an overview of our proposed network architecture for real-time detection of organs from medical images. The network architecture is designed to include prior protocol and contextual organ information for improved performance both in terms of accuracy and speed. The framework builds on state-of-the-art regional proposal networks (RPN) \cite{akselrod2016region} and Faster R-CNN (Region-based Convolutional Neural Networks) \cite{ren2015faster} (Fig. \ref{fig:rpn_original}), which we briefly overviewed next.
\vspace{-0.1in}
\subsection{Faster R-CNN}
Faster R-CNN is the current state-of-the-art method in real-time object detection and has been used in various computer vision applications such as image recognition, visual understanding, etc. Fig. \ref{fig:rpn_original} provides an overview diagram of Faster R-CNN, which consists of two primary modules: a deep convolutional neural network, the RPN, which provides region proposals for the object location hypotheses, and a detection module that uses the proposed region hypotheses and assigns labels to region proposals. 
\vspace{-0.1in}
\subsubsection{Hypothesis Generation--RPN:}
Recent advances in object detection in computer vision applications are largely driven by the RPN \cite{ren2015faster, akselrod2016region}. RPN takes as input a convolutional feature map of any size and provides a set of rectangular object proposals, each having an \emph{objectness} score that measures the membership of the region to a set of object classes. Any deep-learning architecture can be used to generate the convolutional feature map. Several architectures such as VGG-16 have been tried to generate a convolutional feature maps \cite{ren2015faster}. To generate region proposals, the RPN slides over every location in the convolutional feature map. Subsequently, each sliding window is mapped to a lower-dimensional feature space (512-D for VGG-16) which is fed into two separate fully connected layers: a box-regression layer (\emph{reg}) and a box-classification layer (\emph{cls}). At each location of the sliding window, RPN predicts multiple region proposals (bounding boxes or anchors) at different scales and aspect ratios. Let $k$ denotes the maximum number of proposals at each location. Then the \emph{reg} layer has $4k$ outputs representing the coordinates of the $k$ bounding boxes (proposals) and the \emph{cls} layer provides $2k$ memberships of the bounding box containing the object(s) of interest or not. Hence, for a convolutional feature map of size $W\times H$, there are $WHk$ proposals in total which can be computationally expensive for medical images. 
\vspace{-0.1in}
\subsubsection{Hypothesis Classification:} For object detection, Faster R-CNN adopts the detection classifier presented in \cite{girshick2015fast}. As shown in Fig. \ref{fig:rpn_original}, Faster R-CNN learns a unified network composed of RPN and detection classifier with shared convolutional layers. After the RPN steps, fixed-sized feature maps are extracted for each proposal using the region of interest (ROI) pooling. Finally, fully-connected layers are used to provide a membership score for each possible object class.
\vspace{-0.1in}
\subsection{Novel Contextual Selective Attention Region Proposal Network}
Similar to Faster R-CNN, the proposed framework consists of two modules: a proposal hypothesis generation module and a detection module, as illustrated in Fig. \ref{fig:rpn_ours}.
\subsubsection{Hypothesis Generation--Selective Attention RPN:}
Medical images are acquired under specific protocols with a predetermined pose. This information can therefore be used to constrain the area of the convolutional feature map in which to look for the organ of interest (instead of scanning the entire map using the sliding window approach). In our proposed selective attention RPN, we exploit that prior information to boost the detection performance of Faster R-CNN in terms of speed and accuracy. Specifically, we use a reduced search space, which we denote as $\mathcal{A}$, to avoid generating region proposals in the areas in which the organ of interest is unlikely to be located. The attention region $\mathcal{A}$ can be determined based on the prior statistical information from the training dataset and the acquisition protocol. Furthermore, we estimated the expected size and aspect-ratio of the organ of interest from the training data and population statistics; this results in fewer and more accurate proposals ($k$) at each location. For the application of left and right lung field detection (two separate classes), we denote the feature map of the last convolutional layer $L$ in the VGG-16 architecture as $\phi^L:\left(x,y\right)|\{0\le x\le W-1, 0\le y\le H-1\}$, so $\phi^L_A:\left(x,y\right)|\{\alpha_1\le x\le \alpha_2(W-1), \beta_1\le y\le \beta_2(H-1)\}$, where $\alpha$ and $\beta$ define the boundaries of the restricted space along the horizontal and vertical dimensions. Furthermore, based on population statistics of lung shape (location, scale, aspect-ratio), we are able to reduce the number of proposals to $4$, as opposed to $6$ used in Faster R-CNN. We trained our contextual selective attention RPN end-to-end using back-propagation with stochastic gradient decent optimization \cite{ferguson1982inconsistent}. Unlike Faster R-CNN, the mini-batch in our approach was allowed to arise from multiple images due to the relative standardization of medical imaging acquisition protocols. In order to minimize the bias effects of having more negative proposals (i.e., proposals with no organ of interest), we randomly sampled positive (i.e., proposals with organ of interest) and negative samples at $1:1$ ratios. Network weights were randomly initialized using a Gaussian distribution with zero mean and 0.01 standard deviation. We used the learning rate of $0.001$ with a weight decay of $0.0005$, and momentum of $0.85$.

\subsubsection{Hypothesis Classification: Contextual R-CNN}
Once the region proposals are obtained using the selective attention RPN, we incorporate their co-ordinates $(x, y, w, h)$, normalized by the image size, to train the detection classifier. This information improves the detection accuracy by incorporating organ location information to its appearance (i.e., the convolutional feature map). We named our detection with appended position information, as contextual R-CNN. As demonstrated later, we found that adding this location information to the appearance information of the proposal reduced the training time by $\approx 30\%$. We use approximate \emph{joint training approach} to train our network \cite{ren2015faster}. Specifically, in this approach, the RPN and the contextual R-CNN networks are merged into a single framework during training, as shown in Fig. \ref{fig:rpn_ours}. At each iteration, the forward pass generates region proposal whose co-ordinates are fed to the contextual R-CNN detector. During back-propagation, the loss from both selective attention RPN and contextual R-CNN are combined as explained below. 

\subsubsection{Loss Function}
To train the RPN module, we assigned a binary class label (object/ no object) to each proposal. A positive class label was assigned to proposals with Intersection over Union (IoU) overlap greater than $0.8$ with any ground-truth bounding-box. A negative label was assigned to the proposals with IoU ratios lower than $0.3$ for all the ground-truth bounding-boxes. The proposals with IoU ratios between $0.3$ and $0.8$ were considered neutrals and therefore not used for training. Using these definitions and adopting the approximate joint training approach, we minimized the following objective function:

\begin{footnotesize}
\begin{equation}
L\left(\{p_i\},\{t_i\}\right)=\sum_i{L_\text{cls}(p_i\mathbb{I}_A, p^*_i\mathbb{I}_A)}+\lambda\sum_i{t^*_iL_\text{reg}(t_i\mathbb{I}_A, t^*_i\mathbb{I}_A)},
\label{eq:loss}
\end{equation}
\end{footnotesize}

where $i$ was the proposal index in the mini-batch, $p_i$ was the predicted probability of the $i^\text{th}$ proposal to not be labeled as background. $p^*_i$ was the ground-truth label for the $i^\text{th}$ proposal ($1$ if the proposal was positive, $0$ if negative). $t_i$ was the vector representing the 4 co-ordinates ($x, y, w, h$), $t_i^*$ was the co-ordinate vector associated with the ground-truth bounding-box. The classification loss $L_\text{cls}$ was the logarithmic loss over two classes (object/ background) while the regression loss ($L_\text{reg}$) was the smooth $L_1$ loss defined in \cite{girshick2015fast}. $L_\text{reg}$ was only activated for positive proposals ($p^*_i=1$) and disabled otherwise ($p^*_i=0$). $\mathbb{I}_A$ was the indicator function defining the attention region $A$ (Fig. \ref{fig:rpn_ours}) of the convolutional feature map, which was our search space. The indicator function ensures that the loss function defined in eq. (\ref{eq:loss}) was calculated using the proposals within $A$ only. $\lambda$ was the weighting parameter between the classification and the regression loss. Although, we set $\lambda=10$ empirically, we did not observe larges effects of using different values of $\lambda$, as also reported in \cite{ren2015faster}.

\section{Experiments}
\subsection{Datasets and Reference Standards}
\label{sec:data}
Our experiments were conducted on both publicly available data and datasets acquired in-house using a wide range of devices, age groups, and multiple pulmonary pathologies. We used 247 (age: \begin{small}$58.21\pm 14.02$\end{small} year) publicly available chest radiographs (CXRs) from the Japanese Society of Radiological Technology (JSRT; http://www.jsrt.or.jp) dataset and 108 (age: \begin{small}$45.60\pm 16.98$\end{small} year) from the Belarus Tuberculosis Portal (BTP; http://tuberculosis.by). In addition, after approval from the Internal Review Board, we used 313 (age: \begin{small}$4.75\pm 5.30$\end{small} year) posterior-anterior CXRs were collected at our institution. The JSRT radiographs had dimensions of \begin{small}$2048\times 2048$\end{small} pixels, spatial resolution of \begin{small}$0.17\times 0.17$\end{small} mm/pixel, and digital resolution of 12 bits. BTP images had dimensions of \begin{small}$2248\times 2724$\end{small} pixels, spatial resolution of \begin{small}$0.16\times 0.16$\end{small} mm/pixel, and the digital resolution of 12 bits.  The ground truth labels of the lung field were prepared using the ITK-SNAP interactive software under the supervision of two expert pulmonologists. 

\subsection{Implementation Details}
\label{sec:iDetails}
Training and validation were performed on a single scale, similar to Faster R-CNN. The images were rescaled to a maximum of 600 pixels on the shorter side. The stride in our framework was 16 pixels. We used 2 different scales to calculate the proposals, with bounding boxes having areas of $66^2$ and $150^2$ pixels, and 2 aspect ratios: 1:2 and 3:4. Based on the statistics of CXR imaging protocol (location, scale, and aspect-ratio of the organ-of-interest), the feature map was shrunk by $15\%$ ($\alpha_1$, $\alpha_2$, $\beta_1$, $\beta_2$) from each side resulting in $51\%$ reduction in the sliding window search space. Similar to Faster R-CNN, non-maximum suppression was applied to the proposals based on their \emph{cls} layer scores. Our framework was implemented using Tensorflow with Keras, and trained using a Nvidia Titan X GPU, CUDA 8.0, and CuDNN 6.0.

\subsection{Results}
\begin{figure}
\begin{subfigure}[b]{0.13\textwidth}
\includegraphics[width=\textwidth]{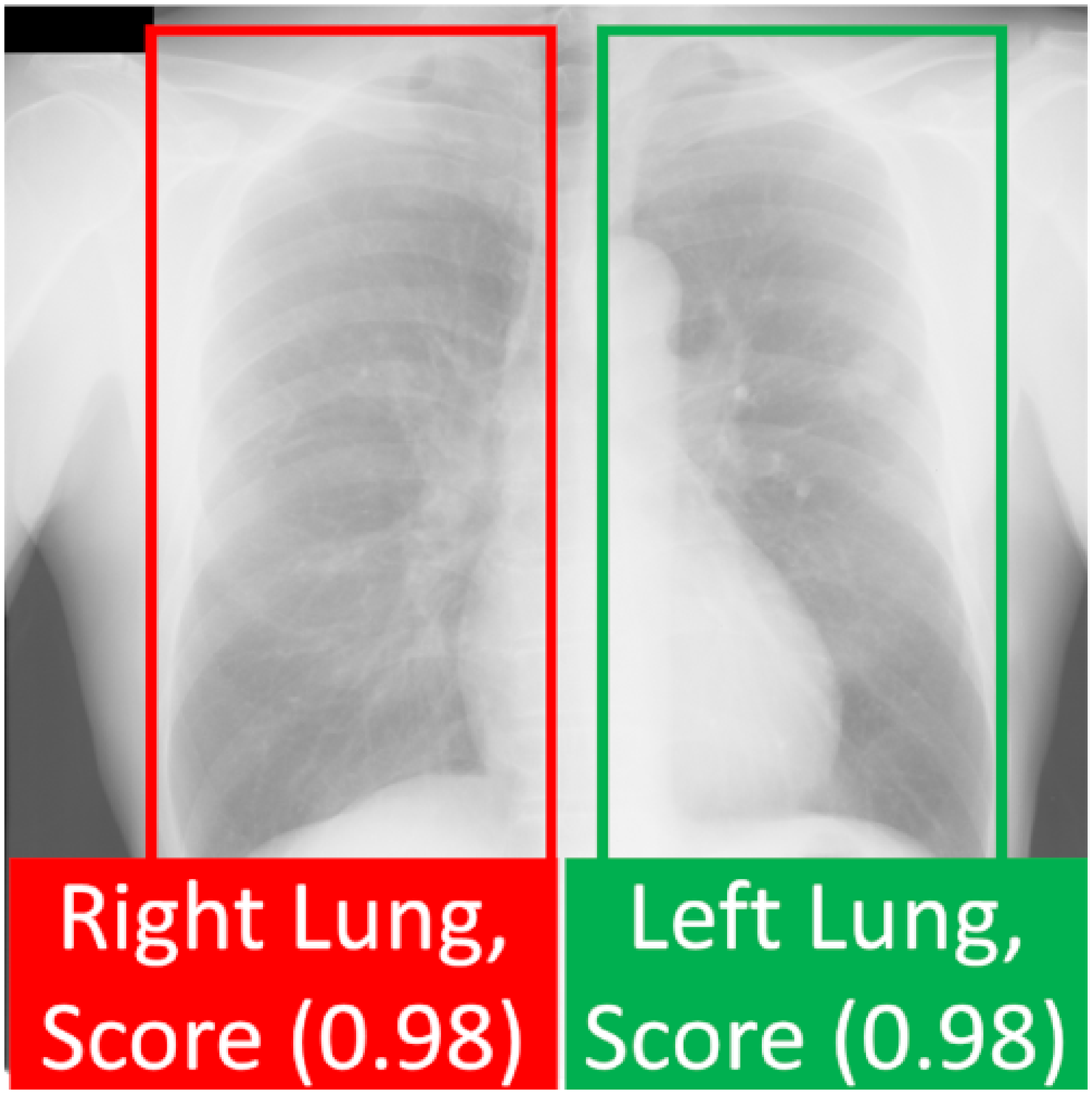}
\caption{}
\end{subfigure}
\begin{subfigure}[b]{0.15\textwidth}
\includegraphics[width=1.1\textwidth]{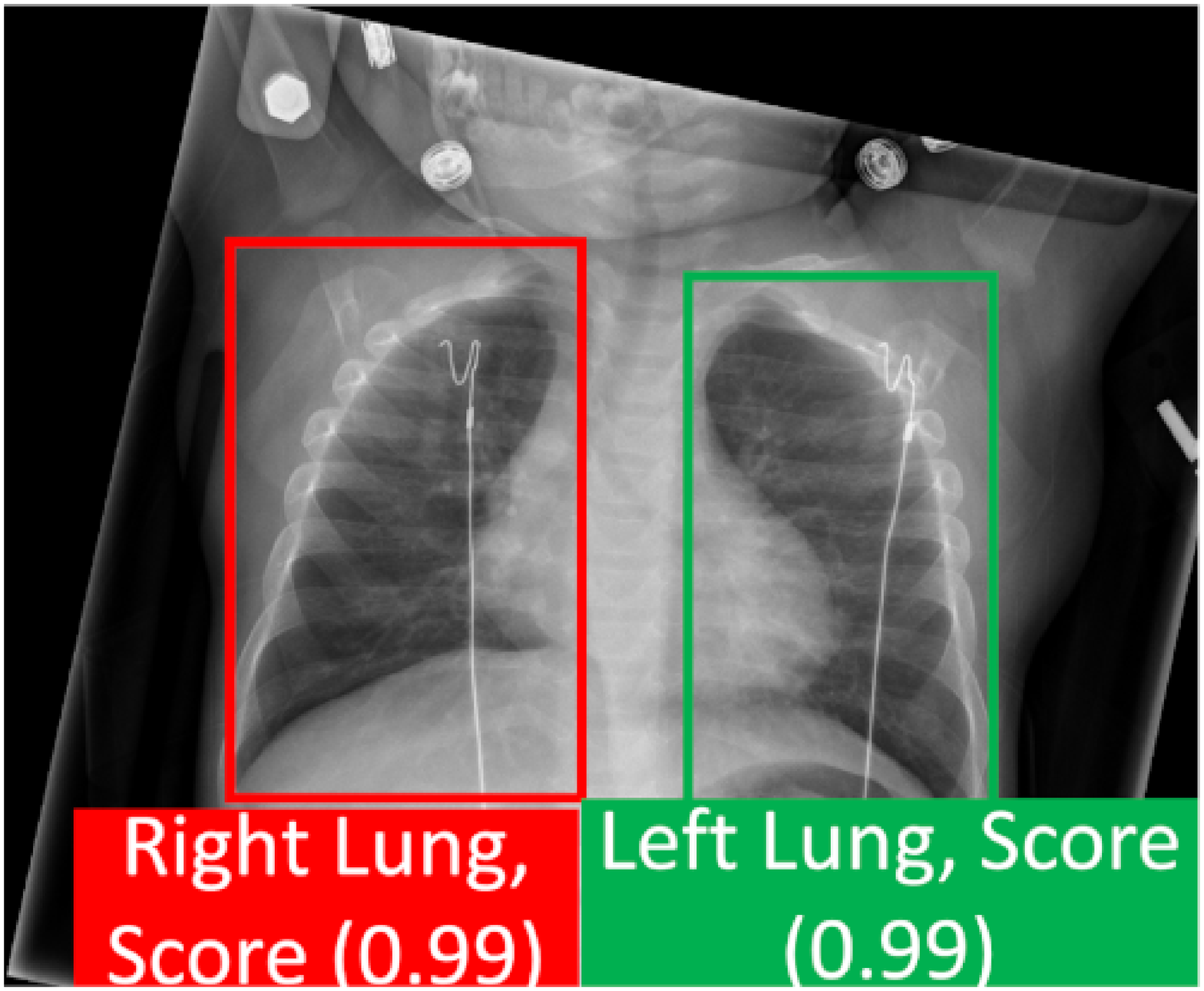}
\caption{}
\end{subfigure}
\begin{subfigure}[b]{0.15\textwidth}
\includegraphics[width=1.1\textwidth]{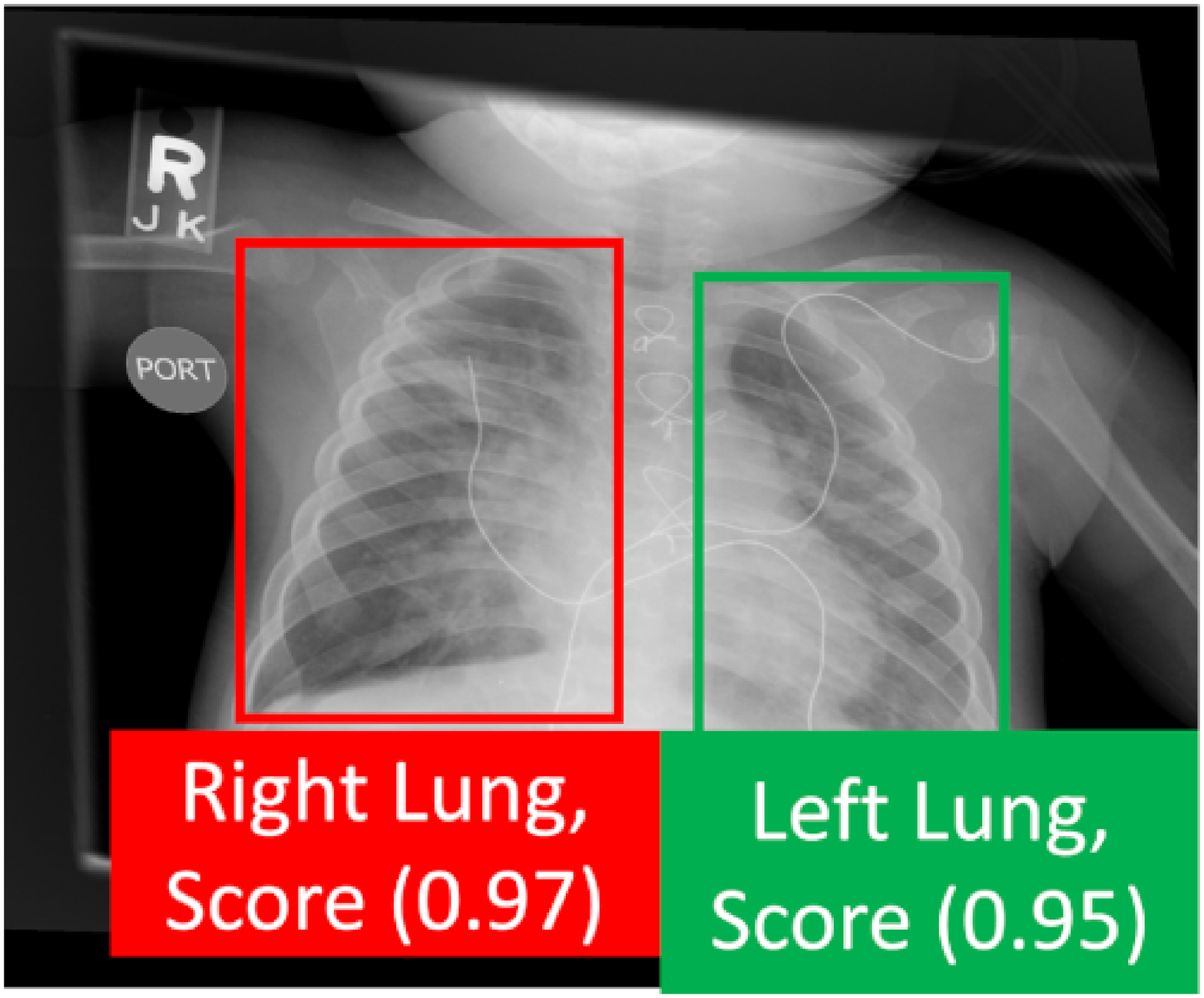}
\caption{}
\end{subfigure}
\caption{\footnotesize{Qualitative results obtained using the proposed method. The bounding-box with detected organ-of-interest (left/ right lung field) and the confidence score by the algorithm is shown.}}
\label{fig:results}
\end{figure}

Table \ref{table:result} compares the performance of the proposed method to the current state-of-the art-- Faster R-CNN \cite{ren2015faster} using the VGG16 architecture-- in terms of the number of proposals, time of execution and detection accuracy. Our framework obtained the Dice score of $0.95\pm0.12$, which was a significant improvement over the Faster R-CNN (p-value$<0.001$; Wilcoxon Rank Sum Test). Interestingly, the detection accuracy of Faster R-CNN was improved by just using the optimal scales and aspect-ratios for the lung field from CXR. These scales and aspect-ratios were empirically calculated in the previous section (Implementation Details). In terms of timing, the proposed framework is $= 27.53\%$ (6.67 fps) faster than Faster R-CNN. Fig. \ref{fig:results} shows our qualitative results.
\begin{table}
\caption{\footnotesize{Performance results comparison of the proposed method with state-of-the-art in terms of the number of proposals, detection accuracy and execution time. The processing time includes non-maximum suppression, pooling, fully-connected, and softmax layers.}\label{table:results}}
%\begin{algorithm*}
%\rule{0.5\textwidth}{0.2pt}
\begin{footnotesize}
\begin{tabular}{p{2.7cm}ccc}
\toprule
\textbf{Method}&\textbf{\# proposals}&\textbf{Dice Score}&\textbf{Timing (sec)}\\
\hline
Faster R-CNN&300&$0.88\pm 0.24$&0.21\\
Faster R-CNN with optimal aspect ratios (2) and scales (2)&300&$0.90\pm0.21$&0.18\\
\midrule
\textbf{Proposed Method}&154&$0.95\pm0.12$&0.15\\
\bottomrule
\end{tabular}
\end{footnotesize}
\label{table:result}
\end{table}
\section{Conclusion}
In this paper, we presented a new and general RPN with contextual attention mechanism to generate region proposals efficiently and accurately for organ detection in medical images. We illustrated the improved performance of our framework on the detection of the lung filed from a cohort of diverse chest radiographs with a variety of pathologies. We increased the classification accuracy by providing organ location information to the classifier. Our results also show that by using critical organ localization and geometric information, the region proposal evaluation speed increases on average by $27.53\%$ to provide real-time results. Finally, our novel architecture improves the overall detection accuracy of the lung field by $7\%$. In our future work we plan to extend our approach to 3D volumetric images where this improvement in time will be extremely important in applications such as query processing.

\begin{footnotesize}
\bibliographystyle{IEEEtran}
\bibliography{myBib}
\end{footnotesize}
\end{document}